%% file: icwsm2021.tex
\def\year{2021}\relax
\documentclass[letterpaper]{article} % DO NOT CHANGE THIS
\usepackage{aaai21}  % DO NOT CHANGE THIS
\usepackage{times}  % DO NOT CHANGE THIS
\usepackage{helvet} % DO NOT CHANGE THIS
\usepackage{courier}  % DO NOT CHANGE THIS
\usepackage[hyphens]{url}  % DO NOT CHANGE THIS
\usepackage{graphicx} % DO NOT CHANGE THIS
\usepackage{natbib}  % DO NOT CHANGE THIS AND DO NOT ADD ANY OPTIONS TO IT
\usepackage{caption} % DO NOT CHANGE THIS AND DO NOT ADD ANY OPTIONS TO IT
\usepackage[switch]{lineno}  %
 \usepackage{MnSymbol,wasysym}

\newcommand{\name}{\texttt{MHA-Meme}}

\input{packages}

\title{{\em {\color{blue}Exercise?} I thought you said {\color{red}`Extra Fries' \smiley{}}}: Leveraging Sentence Demarcations and Multi-hop Attention for Meme Affect Analysis}

\author{Shraman Pramanick, Md Shad Akhtar, Tanmoy Chakraborty \\ {\normalfont Dept. of Computer Science \& Engineering, IIIT-Delhi, India} \\ {\tt \{shramanp,shad.akhtar,tanmoy\}@iiitd.ac.in}}

\begin{document}
% \linenumbers
\maketitle

\begin{abstract}
Today's Internet is awash in memes as they are humorous, satirical, or ironic which make people laugh. According to a survey\footnote{\url{https://www.ypulse.com/article/2019/03/05/3-stats-that-show-what-memes-mean-to-gen-z-millennials/}}, $33\%$ of social media users in age bracket $[13-35]$ send memes every day, whereas more than $50\%$ send every week. Some of these memes spread rapidly within a very short time-frame, and their virality depends on the novelty of their (textual and visual) content. A few of them convey positive messages, such as funny or motivational quotes; while others are meant to mock/hurt someone's feelings through sarcastic or offensive messages.    
Despite the appealing nature of memes and their rapid emergence on social media, effective analysis of memes has not been adequately attempted to the extent it deserves. Recently, in SemEval'20, a pioneering attempt has been made in this direction by organizing a shared task on `Memotion Analysis' (meme emotion analysis). As expected, the competition attracted more than $500$ participants with the final submission of $[23-32]$ systems across three sub-tasks.

In this paper, we attempt to solve the same set of tasks suggested in the SemEval'20-Memotion Analysis competition. We propose a multi-hop attention-based deep neural network framework, called \name, whose prime objective is to leverage the spatial-domain correspondence between the visual modality (an image) and various textual segments to extract fine-grained feature representations for classification. We evaluate \name\ on the `Memotion Analysis' dataset for all three sub-tasks - \textit{sentiment classification}, \textit{affect classification}, and \textit{affect class quantification}. Our comparative study shows state-of-the-art performances of \name\ for all three tasks compared to the top systems that participated in the competition. 
Unlike all the baselines which perform inconsistently across all three tasks, \name\ outperforms baselines in all the tasks on average. Moreover, we validate the generalization of \name\ on another set of manually annotated test samples and observe it to be consistent. Finally, we establish the interpretability of \name.

\end{abstract}

\input{body}

\bibliography{references}
\end{document}

%% file: packages.tex
\usepackage{latexsym}
\usepackage{graphicx}
\usepackage{amsmath}
\usepackage{soul}
\usepackage{subfig}
\usepackage{float}
\usepackage{multirow}
\usepackage{times}
\usepackage{latexsym}
\usepackage{microtype}
\usepackage{array}
\usepackage{arydshln}
\usepackage{colortbl}
\usepackage{soul,framed}

\usepackage{microtype}
\usepackage{enumitem,kantlipsum}
\usepackage{amsmath,amsfonts}
\usepackage{graphicx,accents}
\usepackage{natbib}
\usepackage{lipsum}
\usepackage{tabularx}
\usepackage{graphicx}
\usepackage{tikzpagenodes}
\usepackage{lipsum}
\usepackage{comment}
\usepackage{xcolor}

\usepackage{algorithm}
\usepackage[noend]{algpseudocode}
\makeatletter
\def\algbackskip{\hskip-\ALG@thistlm}
\makeatother

%% file: body.tex
\section{Introduction}
In recent years, Internet memes (or simply memes) have emerged as one of the most frequently circulated entities on social media platforms. In general, memes describe a basic unit of cultural idea or symbol that can be transmitted from one mind to another and inherently, portray the opinion, resentment, fandom, along with the political, psychological, socio-cultural expression of a community. The behavior of memes is also distinctive as memes replicate and mutate, similar to genes in human evolution, during propagation on social media. Despite being so popular and entrenched on online media, it is extremely challenging to leverage automated methods to understand the inherent sentiment/emotion (affect) expressed by memes.

\subsubsection{Motivation:} Memes contain information of both textual and visual modalities; both the modalities often overlap, but are complementary sometimes. Therefore, to analyze the emotion expressed by a meme, both the modalities should be considered simultaneously; and at the same time, monotonous information must be eliminated. Two different memes may sometimes have the same image but can express completely different semantics based on just a few words in the text. For example,  Figures \ref{fig:meme:exm:1} and \ref{fig:meme:exm:5} display exact same image, but vary in four different semantic classes. However, Figures \ref{fig:meme:exm:2} and \ref{fig:meme:exm:3} have rich visual information. Again, Figure \ref{fig:meme:exm:4} has high textual and very little visual content. For such variation of modality information, memes are often hard to classify.

\subsubsection{State-of-the-art:} Annotating memes into different sentiment and affect classes is another challenge, as emotion about memes highly depends upon an individual’s perception of an array of aspects within
society, and could thus vary from one person to another. This phenomenon is known as ``\textit{Subjective Perception Problem}'' \citep{zhao2018affective}, which leads to discrepancy in annotated data. Very recently, emotion analysis of memes has been portrayed as a separate task in SemEval-2020 \cite{chhavi:semeval:2020:memotion}. A large dataset has been released as a part of the shared task. The winner of the shared task achieved  macro-F1 scores of $0.35$, $0.51$, and $0.32$, respectively for three tasks -- sentiment analysis, affect classification and affect quantification (described later). The performance is not significantly high, which demands further work on these tasks.

\input{figure-examples}

\subsubsection{Challenges:}
A meme $M$ is an image consisting of two modalities -- a background image $I$ and some text $T$ at the foreground, referring to a specific situation. The purpose of creating a meme is either to convey humorous or motivational messages, or to mock someone/something in a rhetorical way. The background image itself can have multiple sections usually representing a progressive story line; however, many memes convey their messages through a single image as well. Similarly, the foreground text can also be clustered into multiple sentences/phrases supporting the development or the depiction of the story line. A few example memes are shown in Figure \ref{fig:meme:example}.

Interpreting a meme is a challenging task often because of the implicit world knowledge and overlapping information. Two memes can have same image but can convey different semantics. Similarly, same text corresponds to two different events according to two background images. Moreover, the world knowledge plays a crucial part in establishing a relationship between a meme and its reference situation.  

\subsubsection{Proposed Method:} The current study aims at analyzing the emotion of memes on three dimensions:  \begin{itemize}
    \item \textbf{Sentiment classification --} whether a meme conveys \textit{positive}, \textit{negative}, or \textit{neutral} sentiment;
    \item \textbf{Affect classification --} whether a meme conveys \textit{humorous}, \textit{sarcastic}, \textit{offensive}, \textit{motivational}, or a combination of the four affects;
    \item \textbf{Affect quantification --} what is the quantification of the expressed affect.%\todo{what is b}
\end{itemize}

The foreground texts are critical in extracting the semantic-level information from meme. However, they require special attention depending upon their position in the meme and their reference to a specific region of the background image. Therefore, in the current work, we propose an attentive deep neural network architecture, called \name\ ({\bf M}ulti-{\bf H}op {\bf A}ttention for {\bf Meme} Analysis), to carefully analyze the correspondence between the background image and each text segment at different spatial locations. To do so, at first, we perform OCR (optical character recognition) to extract texts from the meme, and segment them into $l$ sequence of text depending upon their spatial positions. Next, we process each textual segment $t_i$ separately by establishing their correspondence with the background image $I$. We employ two attention frameworks for exploiting the correspondence -- a uni-modal multi-hop attention framework (\textit{aka} {\tt MHA} attention), and  an attention-based multi-modal fusion (\textit{aka} {\tt ATMF}). Subsequently, we combine all $l$ sequence of the processed texts in a multi-hop attention framework for the classifications. The segment-level analysis enables us to leverage various fine-grained features for achieving the final goal.  

\subsubsection{Evaluation:} We evaluate \name\ on the recently published dataset by the SemEval-2020 shared task on `Memotion Analysis' \cite{chhavi:semeval:2020:memotion}. We perform extensive experiments for all three tasks and compare our results with various (in total 15) existing systems on the leader-board of the respective tasks. We observe improved performance against the state-of-the-art systems for all three tasks, in a range of $[+0.5\%, +2.2\%]$. Furthermore, we collected and annotated an additional set of $334$ memes to validate the generalizability of \name.

\subsubsection{Analysis:} We execute extensive ablation experiments with multiple existing baselines to demonstrate the importance of extracting fine-grained features through the segmented text and sophisticated attention mechanism for meme emotion classification. We also employ LIME (Locally Interpretable Model-Agnostic Explanations) \citep{ribeiro2016should} to visualize the importance of relevant features considering it's prediction. 

\subsubsection{Our Contributions:} The major contributions of the paper are as follows: 
\begin{enumerate}
  \item We leverage the correspondence between a meme and its constituent texts depending upon the spatial locations.
  \item We propose an attentive framework that effectively selects and utilizes complementary features from textual and visual modalities to capture multiple aspects of emotions expressed by a meme.    
  \item We report benchmark results for all three tasks -- sentiment classification, affect classification, and affect quantification. 
  \item In order to validate the generalizability of \name, we collected and annotated an additional set of $334$ memes and checked its performance. Furthermore, we establish the interpretability of \name\ using LIME framework.
    
\end{enumerate}
% The rest of the paper is organized in the following fashion:

\subsubsection{Reproducibility:} To reproduce our results, we present detailed hyper-parameter configurations in Table \ref{tab:hyperparameters} and the experiment section. Moreover, the full dataset and code for \name\ is publicly available at 
% \url{https://github.com/ShramanPramanick/MHA-Meme-Affect-Analysis}.
\url{https://github.com/LCS2-IIITD/MHA-MEME}.
  
\section{Related Work}
\label{sec:length}

Multimodal sentiment analysis (MSA) has gained increasing attention in recent years with the surge of multimedia  data on the Internet and social media. Unlike text sentiment analysis \citep{wang2011topic, zadeh2018multi} which is vastly studied, multimodal approaches make use of visual and acoustic modalities in addition to the textual modality, as a valuable source of information for accurately inferring emotion states. One of the key challenges in multi-modal framework is to utilize and fuse the relevant and complementary information for the prediction. \citet{poria2017review} presented an overview of different categories of fusion techniques and potential performance improvements with multiple modalities. 

\subsubsection{Multimodal Fusion:}
There are mainly three types of fusion strategies for MSA -- early, late and hybrid. Early-fusion directly integrates multiple sources of data into a single feature vector and uses a single classifier \citep{ poria2016convolutional, zadeh2017tensor} for prediction from combined feature representation. Such early-fusion techniques can not exploit complementary nature of multiple modalities and often produces large feature vectors with redundancies. On the other hand, late-fusion refers to the aggregation of decisions from multiple sentiment classifiers, each trained on separate modalities \citep{cambria2016affective, cao2016cross}. These schemes are based on the assumption that separate modalities are independent in the feature space, which is not always true in practice, and thus leading to limited performance when multiple modalities tend to be inter-connected. In contrast, hybrid-fusion employs an intermediate shared representation layer by merging multiple modality-specific paths and has been most successful in the literature. \citet{you2016cross} proposed a cross-modality consistent regression (CCR) scheme for joint textual-visual sentiment analysis. In the similar line, \citet{aldeneh2017pooling} compared pooling methods in the context of valence prediction. In another related work, Memory Fusion Network (MFN) \citep{zadeh2018memory} was proposed to associate a cross-view relevance score to each LSTM for MSA. Recently, \citet{akhtar2019multi} introduced context-level inter-modal attention framework for simultaneously predicting sentiment and emotions of an utterance and suggested that multitask learning framework offers improvement over the single-task framework.

\subsubsection{Meme Emotion Analysis:}

In the last few years, the growing ubiquity of memes on social media has been increasingly crucial to understand the opinion of a community. However, interpreting memes is challenging due to its inherent complex cognitive aspects of emotions. Although there has been a lot of work to understand the sentiment of other social media contents \citep{cambria2016affective, yue2019survey}, such as textual or visual opinions, ratings, movie/product reviews, or recommendations, meme sentiment analysis has not been explored much. The pioneering attempt has recently been made by SemEval-2020 Task 8: Memotion Analysis-The Visuo-Lingual Metaphor! \cite{chhavi:semeval:2020:memotion}. The top participants in this challenge have used a wide variety of methods for unimodal feature extraction, such as FFNN, Naive Bayes, ELMo \citep{peters2018deep}, MMBT \citep{rahman2019m}, BERT \citep{devlin2018bert} for textual modality and Inception-ResNet \citep{szegedy2016inception}, Polynet \citep{zhang2017polynet}, DenseNet \citep{huang2017densely} and PNASNet \citep{liu2018progressive} for visual modality. Some of the participants have also employed a modality-specific deep neural ensembles to incorporate visual and textual information. However, none of these methods can guarantee non-redundant features from two highly correlated modalities to capture every aspect of emotions expressed by a meme.

To this end, we are the first to address this ill-posed problem by {\em introducing a multi-hop attention and image encoding filter based sentiment classifier network} which extracts complementary features from both modalities with overlapping information and achieves state-of-the-art performances on Memotion 1.0 dataset. We  demonstrate the significance of spatial information for effective fusion of both modalities. Extensive experiments with our proposed \name\ against various existing baselines reveal the importance of text segmentation and fine-grained feature extraction to capture multiple aspects of emotion for meme affect analysis.%\to
\input{figure-architecture}

\section{Proposed Methodology}
In this section, we describe our proposed system, \name, for  meme analysis. We take a meme as an input and extract the segmented text using Google OCR Vision API\footnote{\url{https://cloud.google.com/vision}}. For each segmented text ($t_i$) and image\footnote{Image remains same for each segment.} ($I$) pair, we encode them through Bidirectional LSTM \cite{hochreiter1997long} and VGG-19 \cite{simonyan2014very} networks, respectively. Further, we fine-tune the image encoding through a text-aware filter. The objective is to restrict the system to extract features from the spatial region of the text in the meme. Subsequent to the unimodal feature extraction step, we employ a novel unimodal multi-hop attention (MHA) framework to attend relevant features from the respective encoding.

For the affect classification and quantification, we learn all four affect dimensions together in a joint framework, and the intuition of multi-hop attention is to extract different set of features for different affect classes. We observe positive effect of the multi-hop attention for the sentiment classification as well, which further suggests that it is capable of extracting diverse features for the downstream tasks. In the next step, we employ an attention-based multi-modal fusion (ATMF) mechanism to leverage the correspondence between the textual and visual modalities. 

We repeat the above procedure for each textual segment, $t_i, \forall i =[1...l]$, and compute its respective multimodal feature representations. Finally, we combine these representations through another multi-hop attention module to attend to relevant segments for the given meme. The context-aware attended representation is then fed to a sequence of fully-connected layers for the prediction. For the affect classification and quantification, our network includes separate fully-connected layer followed by softmax  for each affect class. In subsequent subsections, we discuss these modules in details. Figure \ref{fig:boat1} presents the architecture of \name, and the procedure is summarized in Algorithm \ref{alg:model}. 

\subsection{Unimodal Feature Extraction}
This module extracts  unimodal features for the textual and visual modalities of an Internet meme. 
\subsubsection{Textual Feature Extraction:}
    We employ 200-dimensional GloVe word embedding  \cite{pennington2014glove} for encoding each word of the textual segment. We attain $77.22\%$ word coverage for the underlying dataset, whereas the remaining out-of-vocabulary words (OOVs) are initialized randomly. 
    We train our network in dynamic mode to allow fine-tuning of the embedding layer via backpropagation. 
    
    We use a $2$-layered BiLSTM for the textual feature extraction. It takes $200$ dimensional embedded word vectors as input and maps them to $2u$ dimensional hidden states, where $u$ is the hidden dimension for each unidirectional LSTM. Mathematically, for the textual segment $t_{i} =  (w^1,w^2,....,w^n)$, we compute 
    \begin{eqnarray}
     \overrightarrow{h_t} & = & \overrightarrow{LSTM}(w_t, \overrightarrow{h_{t-1}}) \\
     \overleftarrow{h_t} & = & \overleftarrow{LSTM}(w_t, \overleftarrow{h_{t+1}}) \\
     h_t & = & [\overrightarrow{h_t}, \overleftarrow{h_t}]
    \end{eqnarray}
    where, $\overrightarrow{h_t} \in \mathbb{R}^{u}$, $\overleftarrow{h_t} \in \mathbb{R}^{u}$, and $h_t \in \mathbb{R}^{2u}$ are the forward, backward, and the combined hidden representations, respectively. Moreover, $H = (h_1,h_2,....,h_n) \in \mathbb{R}^{n\times2u}$. We use $u=256$ in our experiments to match with visual feature dimension.
    
    \subsubsection{Image Encoding Filter:}
    We employ VGG-19\footnote{\url{http://www.robots.ox.ac.uk/˜vgg/research/very_deep}} pretrained on ImageNet \citep{simonyan2014very} to extract local feature maps for the images. Specifically, we extract $7\times 7\times 512$ feature maps from the last pooling layer (pool5) of VGG-19 and reshape it into a matrix $F = (f_1,..,f_m), m$~$=$~$49$, where each $f_k\in \mathbb{R}^{512}$ corresponds to a local spatial region.
    
    In general, visual modality contains complementary information from the textual modality. However, while dealing with meme images, we need to ensure that the OCR-extracted text and the text in the image do not establish a direct correspondence. Therefore, to restrict our model to focus aggressively on the textual parts of the image and extract redundant visual features, we aim to filter the redundant visual features in correspondence with the textual utterance  \citep{lee2018stacked, liu2019dynamic}.
    
    Given $H\in \mathbb{R}^{n\times 512}$ and $F\in \mathbb{R}^{m\times 512}$, the image encoding filter mechanism begins with defining an affinity matrix $C\in \mathbb{R}^{n\times m}$, whose element $c_{ij}$ denotes the similarity between the feature vector pair, $h_i\in \mathbb{R}^{512}$ and $f_i\in \mathbb{R}^{512}$:
    \begin{eqnarray}
     C & = & tanh(H\mathbf{W}^bF^\top)
     \end{eqnarray}
     where, $\mathbf{W}^b\in \mathbb{R}^{512\times 512}$ is a correlation matrix to be learned during training.
     
     Subsequently, we compute a normalized weight $\alpha^{h}_{ij}$ to denote the relevance of the $i^{th}$ word to the spatial region $j$. Hence, the weighted summation of all word representation can be represented as,  
    \begin{equation}
     a_j^h  =  \sum\limits_{i=1}^n \alpha^{h}_{ij} h_i, \text{ where    }
     \alpha^{h}_{ij}  =  exp(c_{ij})/\sum\limits_{i=1}^n exp(c_{ij})
    \end{equation}
     
     Since our goal is to emphasize the dissimilar features between the textual and visual modalities and to determine the importance of each image region given the textual utterance, we define the relevance matrix $R(f_i,a_j^h)$ as cosine distance between the attended sentence vector $a_j^h$ and image region feature $f_i$ --
     \begin{eqnarray}
     R(f_i,a_j^h) & = & 1 - \dfrac{f_i^\top \cdot a_j^h}{\| f_i\| \| a_j^h\| }
     \end{eqnarray}
     
     Finally, the weighted summation of all regions gives the modified image representation $U$ computed as,
     \begin{eqnarray}
     U & = & \sum\limits_{j=1}^m R(f_i,a_j^h) \cdot f_i
     \end{eqnarray}
     where, $R(f_i,a_j^h)$ acts as a filter for the image encoding $f_i$.

\subsection{Multi-hop Attention for Unimodal Feature Representation}
Every word in a sentence has a specific role in the semantic space; however, some words can have a higher relevance to a particular task. Hence, semantic attention mechanism has been proven to be extremely beneficial for many NLP tasks \citep{bahdanau2014neural, wang2016attention}. Similarly, some visual regions are often more informative to represent certain emotions, and the literature \citep{lu2017knowing, you2017visual} supports visual attention mechanism to be an effective solution.   

Traditional self-attention mechanism typically focuses on a specific component of a sentence or image, reflecting only one aspect of the semantics. However, a meme can express multiple emotions together, e.g., a meme can be humorous as well as sarcastic. Therefore, we hypothesize that to cater to different tasks in a joint-framework, multiple hops of attention mechanism would attend to diverse and task-specific relevant features, i.e., one hop would attend to features relevant for the humor classification, while the other would focus on the sarcasm detection. 

Following our intuition, we compute multi-hop attention as follows: Given a unimodal feature matrix (e.g., the textual features $H\in \mathbb{R}^{n\times 512}$ or the visual features  $U\in \mathbb{R}^{m\times 512}$), we aim to extract $k$ distinct set of relevant features corresponding to the input representation, where $k$ is a hyper-parameter. The multi-hop attention mechanism takes $H$ as an input, and outputs an attention weight matrix $A \in \mathbb{R}^{k\times n}$ respective to the textual representation. Similarly, we compute attention weight matrix $B \in \mathbb{R}^{k\times m}$ for the visual representation $U$.
\begin{eqnarray}
    A & = & softmax(\mathbf{W_{h2}}\ \ tanh(\mathbf{W_{h1}} H^\top))  \\
    B & = & softmax(\mathbf{W_{u2}}\ \ tanh(\mathbf{W_{u1}} U^\top))
\end{eqnarray}
where, $\mathbf{W_{h1}},\mathbf{W_{u1}}\in \mathbb{R}^{d\times 512}$ and  $\mathbf{W_{h2}},\mathbf{W_{u2}}\in \mathbb{R}^{k\times d}$ are parameter matrices to be learned during training. The $softmax(\cdot)$ is performed along the second dimension of its input, and $d$ is a hyper-parameter we can set arbitrarily. 

The resulting embedding matrices $M$ and $N$ respectively for textual and visual modalities  are computed using their respective attention weight matrices and unattended features.
\begin{eqnarray}
    M & = & A \otimes H\\
    N & = & B \otimes U
\end{eqnarray}
where, $M, N\in \mathbb{R}^{k\times 512}$ are self-attended features for the textual and visual modalities. 

\input{algorithm}
\input{table-dataset}

\subsection{Attention-based Multi-modal Fusion (ATMF)}
ATMF utilizes an attention based mechanism to fuse the unimodal features from textual and visual modalities. Since same modality may have different contribution for different utterances and different meme samples, the attention-based multi-modal fusion network shows much better classification performance over concatenation and neural fusion techniques. 
Extending the attentive fusion network of \citep{gu2018hybrid}, our ATMF module consists of two major parts -- modality attention generation and weighted feature concatenation. In the first part, we use a sequence of dense layers followed by a softmax layer to generate the weighted scores for two given modalities. Each successive dense layer has smaller dimension than its previous one, thus forming a tower like architecture. 
\begin{eqnarray}
    [s_t, s_v] & = & softmax(Dense(M, N))
\end{eqnarray}
where, $[s_t,s_v]$ are the attention scores for the textual and visual modalities, respectively. 

In the second part, the original unimodal features are weighted using their respective attention scores and concatenated together; and $(1 + s_t)M$ and $(1 + s_v)N$ denote residual+attended vectors for the textual and visual modalities respectively.
\begin{eqnarray}
   P_F & = & tanh(\mathbf{W_F}\cdot[(1+s_t)M, (1+s_v)N]
\end{eqnarray}
where, $W_F\in \mathbb{R}^{512\times 512}$ is a learnable parameter.

Moreover, we incorporate an additional attention layer specifically to reduce the effect of repetitive features captured in MHA for a short text segment (e.g., a single-word text segment). In the absence of sufficient words in a text segment, the multi-hop attention module would compute repetitive features for $k$ hops in both textual and visual modalities; thus we hypothesize that the inclusion of an additional attention layer would address this problem. Therefore, we compute the final segment-level multi-modal representation $x \in \mathbb{R}^{512}$ as follows:
\begin{eqnarray}
   \gamma_f & = & softmax(\mathbf{w_f}^\top \cdot P_F)\\
   x & = & [(1+s_t)M, (1+s_v)N] \cdot \gamma_f^\top
\end{eqnarray}
where, $w_f\in \mathbb{R}^{512}$ is a learnable parameter.

\subsection{Context-aware Classification}
It is the final stage of our network where we leverage the contextual multimodal representation of each textual-segment for the classification. However, the sentiment or the affective dimension sometime relates to a few specific textual segments. Thus, we employ the multi-hop attention module at the meme-level. The objective is to highlight diverse features respective to the underlying tasks.      

Let the matrix $X = [x_1,x_2,....,x_{l}] \in \mathbb{R}^{512\times l}$ contains the multimodal features for each textual segment, where $l$ is the number of segments in a meme and $x_i\in \mathbb{R}^{512}$ is the multimodal feature representation for $i^{th}$ segment. 
Finally, the multi-hop segment-level representation $X^*$ is flattened and forwarded to the softmax layers the classification (i.e., one softmax layer for the sentiment classification and four softmax layers each for the affect classification and quantification tasks.
\begin{eqnarray}
    Z & = & softmax(X^* \cdot \mathbf{W_{soft}} + \mathbf{b_{soft}})\\
    \hat{y} & = & \underset{j}{\arg\max}(Z[j]) \;\; \forall j\in class
\end{eqnarray}
where,  $\mathbf{W_{soft}}$ and $\mathbf{b_{soft}}$ are learnable weights, and $\hat{y}$ is the final predicted class.
 
\section{Experiments}

In this section, we present details of the dataset, data pre-processing steps and training methodologies. 

\subsection{Dataset}
We conduct our evaluations on Memotion\footnote{ \url{http://alt.qcri.org/semeval2020/index.php?id=tasks}} 1.0 dataset \citep{chhavi:semeval:2020:memotion} (dubbed as {\bf Memotion} dataset) which was recently released as part of the SemEval-2020 shared task on `Memotion Analysis.' This dataset consists of $8,480$ manually annotated English memes from 52 unique and globally popular categories, e.g., \textit{Hillary Clinton}, \textit{Donald Trump}, \textit{Minions}, \textit{Baby godfather}, etc. The dataset was annotated through Amazon Mechanical Turk, where entities were annotated into three sentiment classes (i.e., \textit{positive}, \textit{neutral}, and \textit{negative}) and four different affect classes (i.e.,  \textit{Humorous}, \textit{Sarcasm}, \textit{Offensive}, and \textit{Motivation}). The dataset also has quantification scores to which a particular affect is expressed. To address the ``\textit{Subjective Perception Problem}
'' \citep{zhao2018affective}, the annotation process was performed multiple times, and the final annotations were adjudicated based on majority voting.
A brief statistics of the dataset is shown in Table \ref{tab:dataset}. Each instance of the dataset consists of an image, representing a meme, and an unsegmented OCR-extracted text (i.e., an {\tt OCR} text in a meme is represented as a sequence of words without any segment demarcations). The original dataset has $6,992$ and $1,879$ memes in the training and test (Test$_A$) sets, respectively.

Since we need segmented text ({\tt OCR}$_{Seg}$) for building our model, we employed Google OCR Vision API to extract textual segments. However, during segmentation, we encountered some alignment issues with $\sim 500$ memes in the training set. We manually corrected most of the segmentation issues.

From the resultant $6,601$ memes in the training set, we extract $14,032$ textual segments; whereas, $4,184$ textual segments were extracted from $1,879$ meme in the test set. The distributions of sentiment and affect classes for both train and test sets are mentioned in Table \ref{tab:dataset}. For the sentiment classification, we classify each meme into `\textit{positive}', `\textit{neutral}', or 
`\textit{negative}' classes. The affect classification is a multi-label problem, where a meme can belong to more than one class, i.e., any combination of the `\textit{humor}', `\textit{sarcasm}', `\textit{offensive}', and `\textit{motivational}' affects is possible. The affect class quantification is a fine-grained classification task, which determines the extent of the expressed affects. The quantification of 
`\textit{humor}' is `\textit{not funny}', `\textit{funny}', `\textit{very funny}', and `\textit{hilarious}'; whereas for sarcasm, it is `\textit{not sarcastic}',  `\textit{general}', `\textit{twisted meaning}', and `\textit{very twisted}'. Similarly for offensive, the quantification labels are `\textit{not offensive}', `\textit{slightly offensive}', `\textit{very offensive}', and `\textit{hateful offensive}
'. In contrast, the `\textit{motivation}' affect has only two extents, i.e., `\textit{not motivational}' and `\textit{motivational}'.

We also collected and annotated an additional set of 334 memes, called Test$_B$, to validate the generalization of \name. For annotation, we followed the guidelines as laid out by \citet{chhavi:semeval:2020:memotion}. Table \ref{tab:dataset} shows the statistics of Test$_B$.

\subsection{Training}
We train \name\ using Pytorch framework on a NVIDIA Tesla T4 GPU, with 16 GB dedicated memory, with CUDA-10 and cuDNN-11 installed. We employ pre-trained GloVe \cite{pennington2014glove} Twitter embedding model\footnote{\url{https://nlp.stanford.edu/projects/glove/}} and fine-tune them during training. The network is randomly initialized with a zero-mean Gaussian distribution with standard deviation 0.02.  

From Table \ref{tab:dataset}, we can observe label imbalance problem for both sentiment ([\textit{positive} and \textit{negative}] vs. \textit{neutral}) and affect ([\textit{humor}, \textit{sarcasm}, and \textit{offense}] vs. \textit{motivational}) classification tasks.
To minimize the effect of label imbalance in loss calculation, we assign larger weights for minority classes. We train our models using Adam \cite{kingma2014adam} optimizer and negative log-likelihood (NLL) loss as the objective function. In Table \ref{tab:hyperparameters}, we furnish the details of hyper-parameters used for the training.

\input{table-hyperparameters}

\input{table-ablation}
\subsection{Baselines}
We compare the performance of  \name\  with various existing models on the leader-board of three separate tasks in SemEval-2020. A brief description of each of these baselines is given below.

\begin{itemize}[leftmargin= 0.1in]
\item \textbf{Guoym:} \citet{guo-etal-2020-guoym} trained five base classifiers to utilize five different types of data representation and combined their outputs  through data-based and feature-based ensemble methods. Their system performed consistently well across three different sub-tasks and ranked $2^{nd}$, $2^{nd}$, and $1^{st}$ in sentiment classification, affect classification, and affect quantification, respectively, in the SemEval’20-Memotion Analysis competition.

\item \textbf{Vkeswani IITK:} \citet{keswani2020iitk} used a simple feed-forward neural network with Word2vec embedding for all the three subtasks. They utilized domain knowledge to improve the classification of context-specific memes.

\item \textbf{George.Vlad Eduardgzaharia UPB:} \citet{vlad-etal-2020-upb} developed a multimodal multi-task learning architecture that combines pre-trained ALBERT for text encoding with pre-trained VGG16 for image representation. They fused  two modalities by simple feature concatenation.

\item \textbf{HonoMi Hitachi:} \citet{morishita-etal-2020-hitachi-semeval-2020} fine-tuned four pre-trained visual (Inception-ResNet, PolyNet, SENet, and PNASNet) and textual (i.e., BERT, GPT-2, Transformer-XL, and XLNet) models, and combined them to  capture the cross-modal correlations between the textual and visual modalities effectively. 

\item \textbf{Mayukh Memebusters:} \citet{sharma-etal-2020-memebusters} used transfer learning for image and text feature extraction. They employed attention-based recurrent (LSTM and GRU) architectures for the final prediction.

\item \textbf{Gundapu Sunil:} \citet{walinska-potoniec-2020-urszula} extracted textual and visual features using LSTM with GloVe word embeddings and pre-trained VGG16, respectively, and fused them for the classification. 

\item \textbf{Souvik Mishra Kraken:} \citet{gupta-etal-2020-bennettnlp} proposed a hybrid neural Naïve-Bayes, Support Vector Machine and Logistic Regression  to solve the multimodal classification problem.

\item \textbf{Nowshed CSECU KDE MA:} \citet{chy-etal-2020-csecu} introduced a convolution and BiLSTM based attentive framework to jointly learn visual and textual features from an input meme.

\item \textbf{Prhlt upv:} \citet{de-la-pena-sarracen-etal-2020-prhlt} used pre-trained BERT to extract textual features and pre-trained VGG19 to extract visual features, and then combined  two modalities using a simple concatenation-based fusion technique. 

\item \textbf{Xiaoyu:} \citet{guo-etal-2020-nuaa} proposed a system which is similar to Prhlt upv, except they found DenseNet outperforming ResNet when used for visual feature extraction. 

\item \textbf{Aihaihara}, \textbf{Saradhix Fermi}, \textbf{Hg}, \textbf{Sourya Diptadas} and
\textbf{Jy930 Rippleai}\footnote{The teams `Aihaihara', `Saradhix Fermi', `Hg', `Sourya Diptadas' and `Jy930 Rippleai' participated in the SemEval’20-Memotion Analysis competition, but did not submit their system description paper in the workshop proceeding. We used their reported scores as outlined in \citet{chhavi:semeval:2020:memotion}.}
\end{itemize}

All the other participants in the SemEval-2020 competition used similar multi-modal deep neural models; however, the difference in hyper-parameters distinguishes among their performance. 
Moreover, these systems did not regard the textual segmentation in their architectures.
To this end, we are the first to analyze the interaction between visual and textual modalities at a fine-grained level, i.e., for each textual segment in a meme, we extract the fine-level feature representation in correspondence with the image. Table \ref{tab:comparison} shows a performance comparison of  \name\ with 15 different baselines taken from SemEval-2021.

\section{Experimental Results}
In this section, we report experimental results for our proposed model, \name, and its variants for all three tasks. At the end, we also present comparative analysis against various baselines which include some of the best systems (current state-of-the-art) submitted as part of the `Memotion Analysis' shared task in SemEval-2021 \cite{chhavi:semeval:2020:memotion}.

For evaluation, we adopt the official evaluation metric of macro-F1 score and perform all analyses and comparisons considering macro-F1 only. Moreover, we report results on both Test$_A$ and Test$_B$ sets. We additionally report micro-F1 during the ablation study as well. 

\subsection{Unimodal Evaluation}
We present our obtained results in Table \ref{tab:ablation}. At first, we experiment with unimodal inputs, i.e., we train separate models for the textual and visual representations. For the textual modality, we employ both BiLSTM and BERT \cite{devlin2018bert} for encoding the text. On the unsegmented {\tt OCR} extracted text, we obtain macro-F1 scores of $0.338$ and $0.336$ in sentiment classification for the BiLSTM and BERT variants, respectively. Similarly for the affect classification and quantification, we yield $0.421$ and $0.302$ macro-F1 with BiLSTM and $0.422$ and $0.295$ macro-F1 with BERT. 
\input{table-comparison}

Subsequently, we include OCR text segments, {\tt OCR}$_{Seg}$, into our model and observe a performance gain in macro-F1 score for all tasks. The BiLSTM variant reports macro-F1 scores of $0.352$ ($+1.4\%$), $0.475$ ($+5.4\%$), and $0.319$ ($+1.7\%$) for the three tasks, respectively. Similarly, we obtain performance improvement of $+1.3\%$, $+5.0\%$, and $+1.1\%$ with BERT. The performance can be easily attributed to the textual segmentation; thus it supports our hypothesis that the different textual segments reflect different semantics and should be processed separately to extract the fine-grained features. We also observe that the performance of BERT, though comparable, is on the lower side as compared to BiLSTM with both segmented and unsegmented inputs. It could be because of shorter segmented text - for which an LSTM performs very well to capture the short-term dependencies. Since BERT is a contextual word embedding model, the lack of context could be a reason for its inferiority. Moreover, in case of memes, the text is often noisy and grammatically incorrect. Instead of using full sentences, memes repeatedly use sentence segments, which are uncommon in structured language resources. 

For visual modality, we employ three widely-used pre-trained image feature extractors -- InceptionV3 \cite{szegedy2016rethinking}, VGG-16 and VGG-19 \cite{simonyan2014very} networks, in the current work. With VGG-19, our model yields macro-F1 scores of $0.325$, $0.413$, and $0.292$ for the sentiment classification, affect classification, and affect quantification, respectively. Compared to the other two tasks, we observe marginally better results with VGG-19, ranging in $0.2-0.5\%$ performance improvement. Therefore, we choose to use VGG-19 encoding as the visual feature extraction for all other experiments.

\subsection{Bi-modal Evaluation}
Finally, we combine the two available modalities in a single system and learn the multimodal interactions for the underlying tasks. Similar to the unimodal case, we experiment with both BiLSTM and BERT variants for the textual encoding. Subsequently, we fuse these encodings with VGG-19 visual encoding in \name. The resultant BERT variant reports macro-F1 scores of $0.356$, $0.508$, and $0.325$ for sentiment classification, affect classification, and affect quantification, respectively. For the same setup, BiLSTM-based \name\ obtains $0.376$, $0.523$, and $0.332$ macro-F1 scores, respectively. We observe macro-F1 improvements of $+2.0\%$, $+1.5\%$, and $+0.8\%$ against the BERT-based system for the three tasks, respectively. Therefore, we prefer BiLSTM encoding in our proposed \name.  

Furthermore, the bi-modal system obtains $+2.4\%$, $+4.8\%$, and $+1.3\%$ improvements against the best unimodal scores for the three tasks, respectively. As the complementary and diverse information are incorporated in a single network, improvements in bi-modal setup are not surprising. Moreover, we observe higher influence of the textual modality on the overall performance. We comprehend that this result is expected as meme text contains richer information than the meme image. For example, a single background image is often used for multiple memes to describe the situation; however, the semantics of the meme is often established through text.

For each case, we show the results obtained on Test$_B$ in Table \ref{tab:ablation}. Similar to Test$_A$, we obtain better results with multi-modal inputs compared to both text and image unimodal inputs. Moreover, the BiLSTM variant yields the best results on Test$_B$ as compared to BERT variant on bi-modal input. Thus, we argue that \name\ not only performs better on Test$_A$, but also generalizes well on unseen random samples in Test$_B$ and is consistent across the two test sets.

\subsection{Comparative Study}
For the comparative study, we pitch our proposed  \name\ against the best performing systems reported by the `Memotion Analysis' shared task \cite{chhavi:semeval:2020:memotion}. We take the official scores of these state-of-the-art and baseline systems from the task-description paper\footnote{\url{https://www.aclweb.org/anthology/2020.semeval-1.99/}} for comparison. We report comparative results in Table \ref{tab:comparison}. We also highlight the official rank of the used baselines as per the shared task portal, the best and the second ranked methods. The first batch of results (after baseline) in Table \ref{tab:comparison} denotes a set of top three ranked systems for the three tasks (on average).

\input{table-multi-hops}

\subsubsection{Sentiment Classification:}
The baseline system of the shared task obtains macro-F1 score of $0.217$ for sentiment classification. The top three submitted systems are `Vkeswani IITK' \citep{keswani2020iitk}, `Gyoum' \citep{guo-etal-2020-guoym}, and `Aihaihara', and their reported macro-F1 scores on the test set are $0.35446$, $0.35197$, and $0.35017$, respectively. The narrow margins among these systems reveal that the meme sentiment analysis is a complex problem and significantly different from the tradition multimodal sentiment analysis. 

In comparison, \name\ yields macro-F1 score of $0.37621$ with the performance improvement of $+2.2\%$ compared to the top system, `Vkeswani IITK' ($0.35446$). We can relate the improvement to the better handling of the fine-grained features through the segmented text and sophisticated attention mechanism.

\subsubsection{Affect Classification:}
For the affect classification, we present class-wise as well as average macro-F1 scores for all systems in Table \ref{tab:comparison}. We observe that \name\ reports new state-of-the-art performance ($0.51936$) for \textit{sarcasm} classification with $+0.4\%$ improvement over the top system, `George.Vlad Eduardgzaharia UPB' \cite{vlad-etal-2020-upb} ($0.51590$).
In the \textit{humor} and \textit{motivation} classification, \name\ obtains second best macro-F1 scores of $0.52670$ and $0.53102$ - a difference of $-0.3\%$ with `Mayukh Memebusters' \citep{sharma-etal-2020-memebusters} ($0.52992$) in \textit{humor} and `Saradhix Fermi' ($0.53411$) in \textit{motivation}. For the offense classification task, we report third best with $0.51731$ macro-F1 score. An important point to observe here is that except `George.Vlad Eduardgzaharia UPB', none of the class-wise best systems (i.e., `Mayukh Memebusters' and `Saradhix Fermi') ranked among top three on average. In comparison, \name\ yields better average macro-F1 score against the top system, `George.Vlad Eduardgzaharia UPB'. Moreover, fine-grained comparison further reveals that in four out of five comparisons (four affects and one average), \name\ performs better than the top system. Also, \name\ records best scores in two comparisons (sarcasm and average), second ranked in two comparisons (humor and motivation), and third ranked in one comparison (offense). In contrast, the top system achieves second rank in three comparisons (sarcasm, offense, and average), fourth rank in one comparison (humor), and sixth rank in one comparison (motivation). Thus, we argue that \name\ is not only the best system on average macro-F1 score but also poses a higher degree of generalization at the task-level as well.    

\subsubsection{Affect Quantification:}
In this task, \name\ performs even better than the affect classification task. We observe that four separate systems, namely `Guoym' (humor), `Vkeswani IITK' (sarcasm) \cite{keswani2020iitk}, `Mayukh Memebusters' (offense), and `Saradhix Fermi' (motivation), report best scores among the submitted systems respective to four affect dimensions (mentioned within parenthesis). Moreover, only one of these systems (`Guoym') ranks in top three as per the average macro-F1 score. In comparison, \name\ yields four best scores (including the official ranking metric, i.e., the average case) and one second best score across all submitted systems.

Across the three tasks, we observe three different systems ranked first in the competition, whereas, \name\ reports state-of-the-art for all three cases. Moreover, for the class-wise affect classification and quantification, five separate systems rank top in eight setups (four each in affect classification and quantification). In comparison, \name\ records state-of-the-art performances in four cases, second best in three cases and third best in one case. 

\subsection{Ablation Analysis}
In this section, we present our analyses for two submodules of \name\ -- we report ablation studies of the multi-hop attention and attention-based multimodal fusion modules. 

The notion of multi-hop attention was introduced by \citet{lin2017structured} to represent the overall semantics of an utterance. This is specifically important for meme sentiment analysis as one single utterance (or textual segment) can often capture multiple emotions together. For example, a meme utterance can be sarcastic and offensive at the same time. To obtain complementary features for different objectives, we incorporate multiple hops of attention mechanism over the same input. For ablation, we show the performance of single-hop and multi-hop in Table \ref{tab:ATMF}. The incorporation of multi-hops yields $\sim2\%$ improvement for different model variants on both Test$_A$ and Test$_B$.  

ATMF employs a hierarchical attention mechanism to amplify the contribution of important modality during fusion. To establish the efficacy of ATMF, we compare our fusion mechanism with two other variants: D-Fusion (Direct fusion) and AT-Fusion \cite{poria2017multi}. We first remove the textual segmentation operation at the input of the network and as a consequence, the whole text in each meme is treated as a single segment. Subsequently, we replace the ATMF layer with simple concatenation (Direct Fusion or D-Fusion) and AT-Fusion layers for the experiments. AT-Fusion, which was originally introduced by \citet{poria2017multi}, takes as an input audio, visual, and textual modalities and outputs an attention score for each modality. We modified the original AT-Fusion to exclude the audio modality. We report the obtained scores on Test$_A$ and Test$_B$ in Table \ref{tab:ATMF}. We can observe superiority of ATMF compared to D-Fusion and AT-Fusion in all experiments.

\input{figure-explainability}
\subsection{Interpretability of \name}

The problem of interpreting complex deep neural models is nontrivial, and at the same time, important for further exploration. Meme understanding is a complicated problem -- it requires a high level of abstraction, background knowledge, and subjectivity, which are sometimes hard to expound even with human reasoning. Besides, Memotion 1.0 dataset complies memes of different topics, subjects, and genres, making the task even more challenging. Based on the prediction results, we observe that the success of the model heavily depends on simultaneous image and text understanding. To comprehend the performance of \name, we use LIME (Locally Interpretable Model-Agnostic Explanations) \citep{ribeiro2016should} -- a consistent model-agnostic explainer to explain the predictions in an explicated and faithful manner. It performs so by learning an interpretable model locally around the prediction.

We choose one sample meme from Test$_{A}$ to visualize the explainability of our proposed model using the LIME framework. The example meme has two textual segments and is correctly classified by \name\ (`positive' sentiment) as shown in Figure \ref{fig:input_meme}. We apply LIME on both image and text individually, and analyze the importance of both modalities in the final classification. The prediction probabilities by \name\ corresponding to positive, neutral, and negative classes are $0.683$, $0.246$, and $0.071$, respectively.  Figure \ref{fig:lime_image} highlights the most contributing super-pixels to positive (green) and neutral (red) classes. As expected, the smiling face of the character, highlighted by green pixels, prominently contributes to the positive class, whereas the red pixels do not reveal any relevant information. Figure \ref{fig:lime_txt} demonstrates the contribution of different words from the meme text to positive and neutral/negative classes. The words `{SMILE}' and `{MORE}' have significant contributions to the positive class; removing these two words drastically reduces the prediction probability of the positive class. Moreover, we also plot the attention weights computed by \name\ in Figure \ref{fig:attention_text} for the same example. Evidently, the word `{SMILE}' has the highest attention weight in the two segments,  supporting the explanations by the LIME framework as well.

Based on the above observations, we can conclude that the proposed \name\ framework extracts and attends to the relevant features from the input representations.

\vspace{-2mm}

\section{Conclusion}
In this paper, we addressed three tasks related to the affect analysis of a meme, namely, \textit{sentiment classification}, \textit{affect classification}, and \textit{affect class quantification}. The {sentiment classification} has three labels $[$\textit{positive}, \textit{negative}, and \textit{neutral}$]$, while the affect classification and quantification have four affect dimensions $[$\textit{humor}, \textit{sarcasm}, \textit{offense}, and \textit{motivation}$]$. We proposed an attention-rich neural framework (called \name) that analyzes the interaction between  visual and textual modalities at fine-granular level, i.e., for each textual segment in a meme, it aims to extract the fine-level feature representation in correspondence with the image. We designed two attention mechanisms - a multi-hop attention module for the unimodal feature extraction and an attention-based multimodal fusion module for computing the interaction between the two modalities. Finally, we combined enriched multimodal representations of all segments via another multi-hop attention layer and forwarded it to the output layer for classification. We evaluated \name\ on the recently released `Memotion Analysis' dataset of SemEval-2020 shared task. We performed extensive experiments for each task and compared the obtained performances against 11 baseline systems (including the winners of the shared task). We observed performance improvements in the range $[+0.5\%, +2.2\%]$ for all three tasks. Furthermore, fine-grained result analysis revealed that \name\ achieved consistently good performances (1st in four, 2nd in three, and 3rd in one) across eight affect dimensions, i.e., four each for {affect classification} and {affect class quantification}. In comparison, baseline systems did not report consistent performance for all the tasks or affect dimensions. 

\section*{Acknowledgments}
The work was partially supported by Wipro Pvt Ltd, India and CAI, IIIT-Delhi. T. Chakraborty would like to thank the support of the Ramanujan Fellowship (SERB).

%% file: figure-examples.tex
\begin{figure*}[ht!]
    \centering
    \subfloat[{\tt [-1,1,1,1,0]}\label{fig:meme:exm:1}]{
    \includegraphics[width=0.18\textwidth, height=0.18\textwidth]{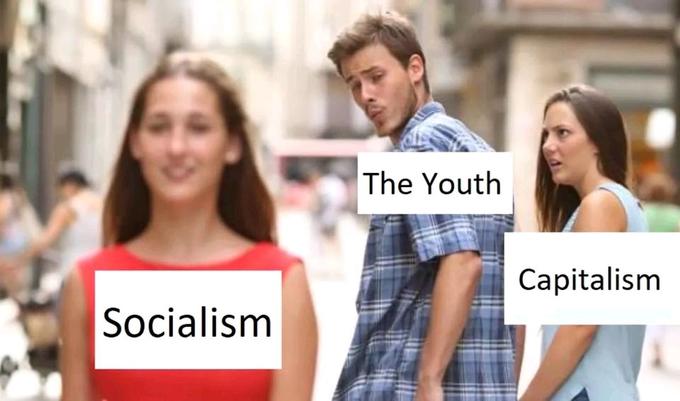}}\hspace{0.5em}
    \subfloat[{\tt [0,0,0,0,0]}\label{fig:meme:exm:5}]{
    \includegraphics[width=0.18\textwidth, height=0.18\textwidth]{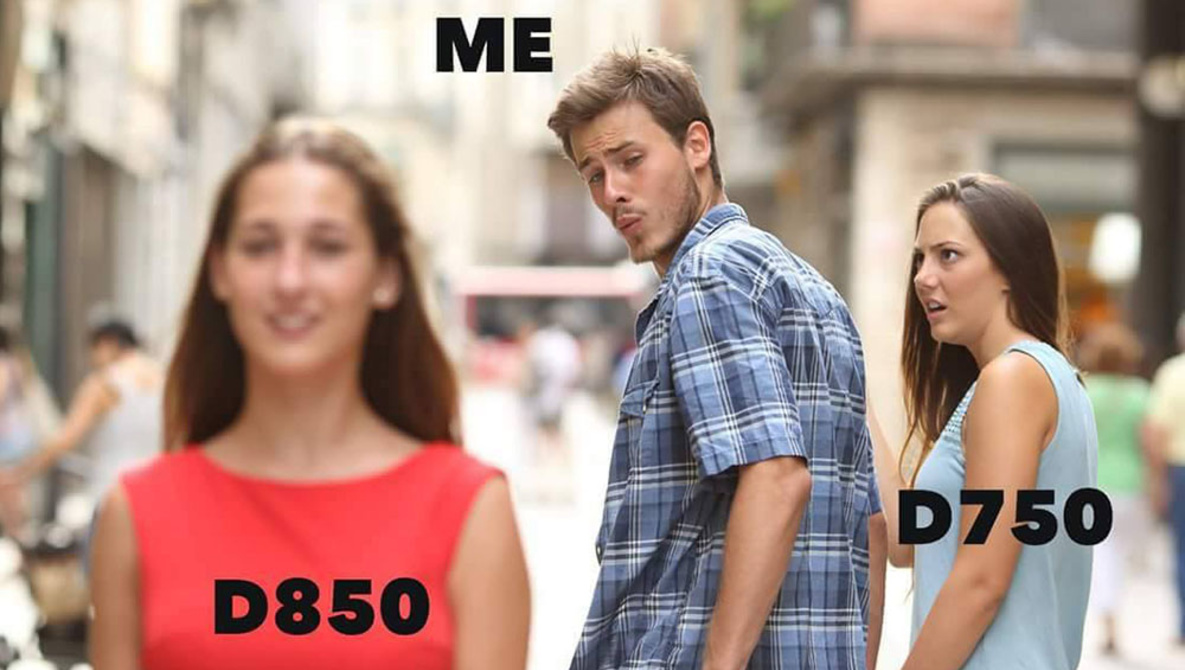}}
    \subfloat[{\tt [1,1,1,1,0]}\label{fig:meme:exm:2}]{
    \includegraphics[width=0.18\textwidth, height=0.18\textwidth]{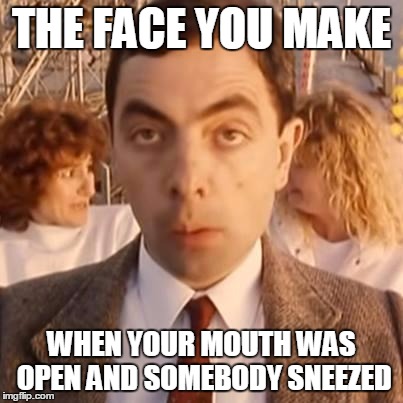}}\hspace{0.5em}
    \subfloat[{\tt [0,1,1,1,1]}\label{fig:meme:exm:3}]{
    \includegraphics[width=0.18\textwidth, height=0.18\textwidth]{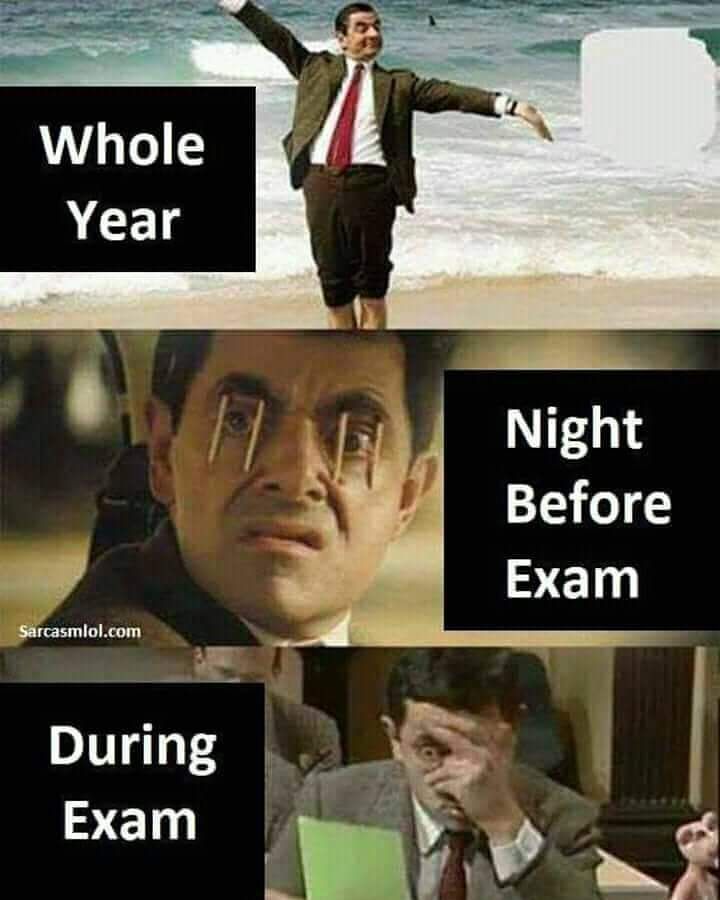}}\hspace{0.5em}
    \subfloat[{\tt [1,1,1,1,0]}\label{fig:meme:exm:4}]{
    \includegraphics[width=0.18\textwidth, height=0.18\textwidth]{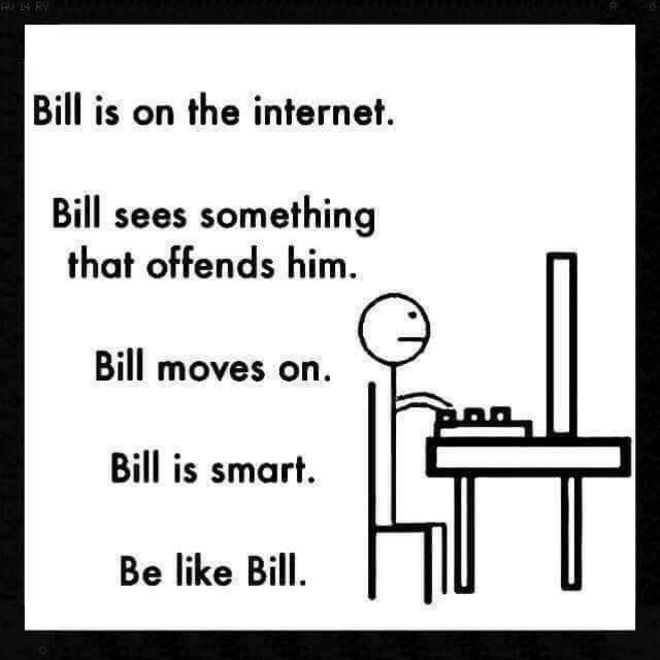}}\hspace{0.5em}
    \caption{A few examples of memes from Memotion Analysis dataset \cite{chhavi:semeval:2020:memotion}. Classification of the memes are in the following format: {\tt [Sentiment, Humor, Sarcasm, Offensive, Motivation]}. For {\tt Sentiment}, \{-1, 0, 1\} corresponds to negative, neutral, positive classes. For other cases,  \{0, 1\} denotes absence and presence of corresponding affect.
    }
    \label{fig:meme:example}
    \vspace{-5mm}
\end{figure*}

%% file: figure-architecture.tex
\begin{figure*}[t!]
\centering
\hspace*{0cm}
  \includegraphics[width=0.85\linewidth]{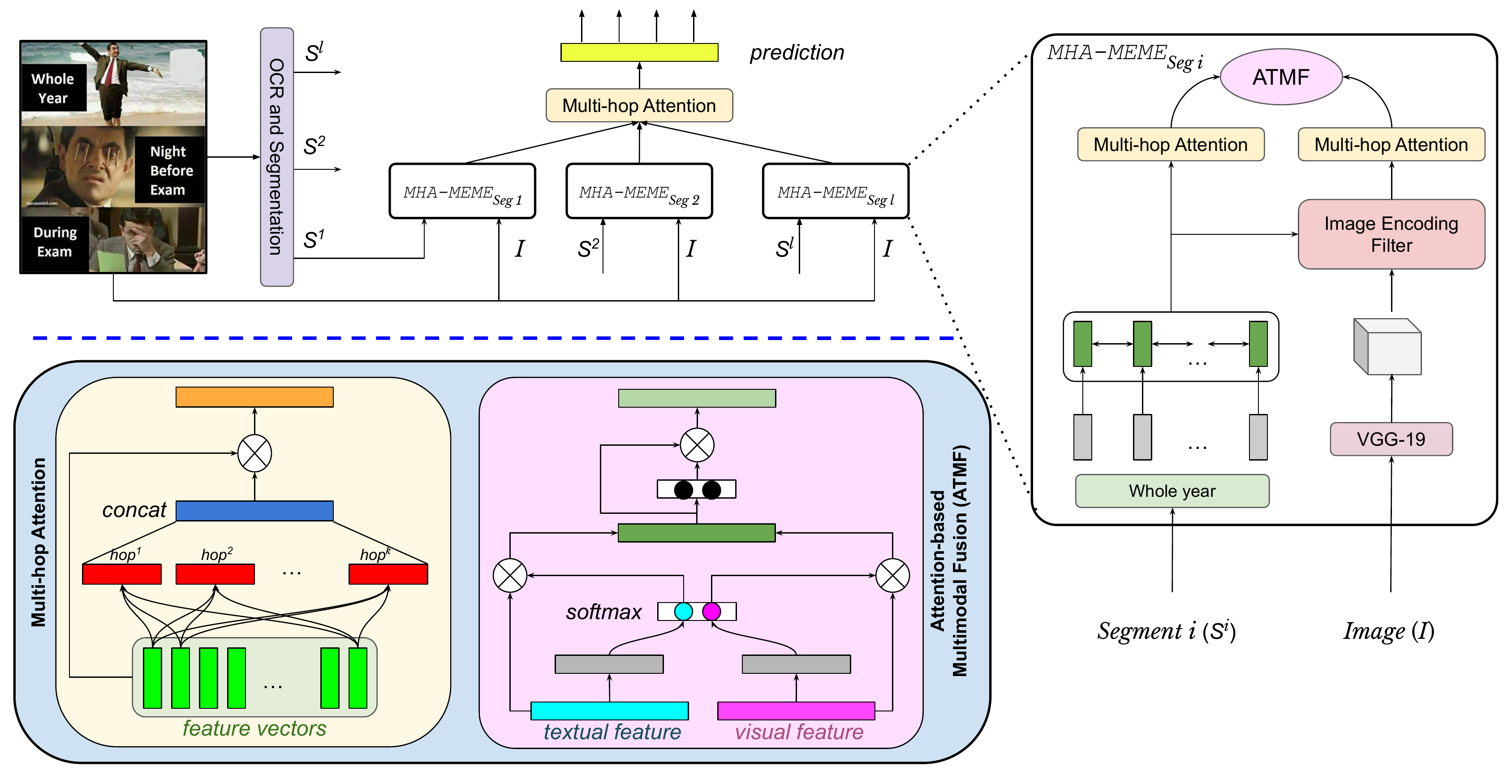}
  \caption{The architecture of the proposed model, \name.}
  \label{fig:boat1}
  \vspace{-5mm}
\end{figure*}

%% file: algorithm.tex
\begin{algorithm}[ht!]
    \caption{\textbf{M}ulti-\textbf{H}op \textbf{A}ttention for \textbf{Meme}}\label{euclid}
    \hspace*{\algorithmicindent} \textbf{Input:} An Internet meme $M$ \\
    \hspace*{\algorithmicindent} \textbf{Output:} Predicted class $\hat y_c$ $\forall$ $c \in \{$affect classes$\}$ 
    \begin{algorithmic}[1]
    \Procedure{MHA-Meme($M$)}{}
    \State $\{t_{1},t_{2},...,t_{n}\}, I$ $\gets$ {\tt OCR} and {\tt Segmentation} ($M$)
    \For{seg in $(1,n)$}  \algorithmiccomment{\#segments}
        \State $u \gets 256$ \algorithmiccomment{LSTM dimension}
        \State $H_{seg}$ $\gets BiLSTM(t_{seg}, u)$
        \State $F_{seg}$ $\gets VGG19_{pool5}(I)$
        \State $F^{filt}_{seg} \gets$ $Image$ $Encoding$ $Filter(H_{seg},F_{seg})$
        \State $M_{seg} \gets$ $Multi$-$hop$ $Attention(H_{seg})$
        \State $N_{seg} \gets$ $Multi$-$hop$ $Attention(F^{filt}_{seg})$
        \State $X_{seg} \gets$ $ATMF(M_{seg}, N_{seg})$ \algorithmiccomment{feature fusion}
     \EndFor
    \State $Z \gets$ $Multi$-$hop Attention(Stack(X_{1},...,X_{n}))$
    \State $\hat y_c \gets$ $Classifier(Z)$ $\forall$ $c \in \{$affect classes$\}$
    \State \Return $\hat y_c$ 
    \EndProcedure
    
    \vspace{0.1cm}
    \Procedure{Image Encoding Filter($H,F$)}{}
    \State $C \gets H\mathbf{W^b}F^\top$ \algorithmiccomment{affinity matrix}
    \State $a^h_j \gets \sum\limits_{i=1}^n((exp(c_{ij})/\sum\limits_{i=1}^n exp(c_{ij}))h_i)$
    \State $R(f_i,a_j^h) \gets 1 - \dfrac{f_i^\top \cdot a_j^h}{\| f_i\| \| a_j^h\| }$  
    \State $U \gets \sum\limits_{j=1}^m R(f_i,a_j^h) \cdot f_i$ 
    \State \Return $U$ 
    \EndProcedure
    
    \vspace{0.1cm}
    \Procedure{Multi-hop Attention($H$)}{}
    \State $A \gets softmax(\mathbf{W_{h2}}\ \ tanh(\mathbf{W_{h1}} H^\top))$
    \State $M \gets A \otimes H$ \algorithmiccomment{attended features}
    \State \Return M 
    
    \EndProcedure
    
    \vspace{0.1cm}
    \Procedure{ATMF($M,N$)}{}
    \State $[s_t, s_v] \gets \sigma(Dense(M, N))$ \algorithmiccomment{$\sigma - softmax$}
    \State $\gamma_f \gets \sigma(\mathbf{w_f}^\top \cdot tanh(\mathbf{W_F}\cdot[(1+s_t)M, (1+s_v)N]))$
    \State $X \gets [(1+s_t)M, (1+s_v)N] \cdot \gamma_f^\top$
    \State \Return X
    
    \EndProcedure
    
    \end{algorithmic}
    \label{alg:model}
\end{algorithm}
% \vspace{-5mm}

%% file: table-dataset.tex
\begin{table*}[t!]
\centering
    
\begin{tabular}{c|c|c|c|c|c|c|c|c|c}
& \multirow{2}{*}{\#Memes} & \multirow{2}{*}{\#Textual segments} & \multicolumn{3}{c|}{Sentiment} & \multicolumn{4}{c}{Affects$^*$} \\ \cline{4-10} 
& & & Pos & Neg & Neu & Humor & Sarcasm & Offense & Motivation \\ \hline 

\hline 
Train & 6601 & 14032 & 3923 & 2092 & 586 & 5082 & 5162 & 4163 & 2406 \\
Test$_A$ & 1879 & 4184 & 1111 & 594 & 173 & 1354 & 1376 & 1101 & 648\\ \hdashline
Test$_B$ & 334 & 748 & 184 & 88 & 62 & 265 & 212 & 244 & 142 \\ \hline  

\hline
\end{tabular}

\caption{Dataset statistics of Memotion Analysis \cite{chhavi:semeval:2020:memotion}. Test$_A$ is the original test set provided by \citet{chhavi:semeval:2020:memotion}. Test$_B$ is developed by us to validate the generalization of \name. Affects$^*$: A meme can belong to $\ge 1$ affective classes.}

\label{tab:dataset}
% \vspace{-6mm}
\end{table*}

%% file: table-hyperparameters.tex
\begin{table}[!t]
\centering % centering table

 \scalebox{0.7}{

\begin{tabular}{p{1.8cm}|p{4cm}|p{2.5cm}|p{2cm}}

\textbf{Module} & \textbf{Hyper-parameter} & \textbf{Notation} & \textbf{Value} \\
\hline

\hline
Feature Extraction & hidden units of BiLSTM & $u$ & $256$ \\
\hline
   & {\tt \#hops} (unimodal) & \multirow{2}{*}{$k$} & $30$ \\
Multi-hop& {\tt \#hops} (multimodal) &  & $10$ \\ \cline{2-4}
Attention& {\tt \#hidden-units }(unimodal) & \multirow{2}{*}{$d$} & $350$\\
& {\tt \#hidden-units} (multimodal) &  & $100$ \\
\hline 
ATMF & {\tt \#neurons} for Dense layers & - & $[256, 64, 8, 1]$ \\
\hline 
\multirow{5}{*}{Training} & {\tt Batch-size} & - & $8$ \\
& {\tt Epochs} & $N$ & $200$ \\ 
& {\tt Optimizer} & - & Adam \\ 
& {\tt Loss} & - & NLL \\ 
& {\tt Learning-rate} & $\alpha$ & $0.005$ \\
& {\tt Learning-rate-decay} & - & $0.0001/10000$ iterations \\
& {\tt Momentum} & - & $0.9$ \\
\hline
\multirow{5}{2cm}{Class weights for imbalanced training data} & sentiment &$[w_{pos}, w_{neu}, w_{neg}]$ & $[1, 1.5, 2]$ \\
& affective - \textit{humor} & $[w_{nonhum}, w_{hum}]$ & $[1.5, 1]$ \\
& affective - \textit{sarcasm} & $[w_{nonsar}, w_{sar}]$ & $[1.5, 1]$ \\
& affective - \textit{offense} & $[w_{nonoff}, w_{off}]$ & $[1.25, 1]$ \\
& affective - \textit{motivation} & $[w_{nonmot}, w_{mot}]$ & $[1, 1.25]$ \\
 \hline
 
 \hline
\end{tabular}}
\caption{Hyper-parameters of \name.}
\label{tab:hyperparameters}
\vspace{-5mm}
\end{table}

%% file: table-ablation.tex
%%%% final version %%%%
\begin{table*}[ht]
\centering % centering table

\setlength{\extrarowheight}{3pt}
\resizebox{0.95\textwidth}{!}
{
\begin{tabular}{c|l|c|c|c|c|c|c|c|c|c|c|c|c}
% \hline

\multirow{3}{*}{\bf Modality} & & \multicolumn{4}{c|}{\bf Sentiment classification} & \multicolumn{4}{c|}{\bf Affect classication (Avg)} & \multicolumn{4}{c}{\bf Affect quantification (Avg)} \\\cline{3-14}

& & \multicolumn{2}{c|}{\bf Macro F1} & \multicolumn{2}{c|}{\bf Micro F1} & \multicolumn{2}{c|}{\bf Macro F1} & \multicolumn{2}{c|}{\bf Micro F1} & \multicolumn{2}{c|}{\bf Macro F1} & \multicolumn{2}{c}{\bf Micro F1}\\ \cline{3-14}

& & \bf Test$_A$ & \textcolor{black}{\bf Test$_B$} & \bf Test$_A$ & \textcolor{black}{\bf Test$_B$} &  \bf Test$_A$ & \textcolor{black}{\bf Test$_B$} &  \bf Test$_A$ & \textcolor{black}{\bf Test$_B$} &  \bf Test$_A$ & \textcolor{black}{\bf Test$_B$} & \bf Test$_A$ & \textcolor{black}{\bf Test$_B$} \\ 

\hline
\hline

\multirow{4}{*}{Text} & BiLSTM - {\tt OCR} & 0.338 & \textcolor{black}{0.373} & 0.509 & \textcolor{black}{0.572} &  0.421 & \textcolor{black}{0.455} &  0.542 & \textcolor{black}{0.570} &  0.302 & \textcolor{black}{0.310} & 0.420 & \textcolor{black}{0.438}\\

& BERT - {\tt OCR} & 0.336 & \textcolor{black}{0.375} & 0.512 & \textcolor{black}{0.570} &  0.422 & \textcolor{black}{0.449} & 0.549 & \textcolor{black}{0.571} &  0.295 & \textcolor{black}{0.298} & 0.395 & \textcolor{black}{0.402}\\

& BiLSTM - {\tt OCR}$_{Seg}$ & 0.352 & \textcolor{black}{0.391} & 0.560 & \textcolor{black}{0.594} & 0.475 & \textcolor{black}{0.490} & 0.570 & \textcolor{black}{0.594} & 0.319 & \textcolor{black}{0.332} & 0.422 & \textcolor{black}{0.442} \\

& BERT - {\tt OCR}$_{Seg}$ &  0.351  & \textcolor{black}{0.384} & 0.538 & \textcolor{black}{0.580} & 0.471 & \textcolor{black}{0.482} &  0.563 & \textcolor{black}{0.581} &  0.311 & \textcolor{black}{0.316}  & 0.418 & \textcolor{black}{0.425}\\ \hline

\multirow{3}{*}{Image} & InceptionV3 &  0.322 & \textcolor{black}{0.358} &  0.516 & \textcolor{black}{0.557} &  0.407 & \textcolor{black}{0.430} &  0.499 & \textcolor{black}{0.525} &  0.288 & \textcolor{black}{0.287} & 0.402 & \textcolor{black}{0.406}\\

& VGG16 &  0.318 & \textcolor{black}{0.355} & 0.521 & \textcolor{black}{0.560} & 0.399 & \textcolor{black}{0.432} &  0.505 & \textcolor{black}{0.532} &  0.286 & \textcolor{black}{0.295} & 0.411 & \textcolor{black}{0.418} \\

& VGG19 & 0.325 & \textcolor{black}{0.367} & 0.525 & \textcolor{black}{0.562} & 0.413 & \textcolor{black}{0.448} &  0.518 & \textcolor{black}{0.550} & 0.292 & \textcolor{black}{0.300} & 0.405 & \textcolor{black}{0.419}\\  \hline

\multirow{2}{*}{Text+Image} & BERT - {\tt OCR}$_{Seg}$ + VGG19 & 0.356 & \textcolor{black}{0.410} & 0.585 & \textcolor{black}{0.624} & 0.508 & \textcolor{black}{0.529} & 0.620 & \textcolor{black}{0.645} & 0.325 & \textcolor{black}{\bf 0.362} & 0.424 & \textcolor{black}{0.435} \\

& BiLSTM - {\tt OCR}$_{Seg}$ + VGG19 & \bf 0.376 & \textcolor{black}{\bf 0.426} & \bf 0.608 & \textcolor{black}{\bf 0.635} & \bf 0.523 & \textcolor{black}{\bf 0.545} & \bf 0.682 & \textcolor{black}{\bf 0.698} & \bf 0.333 & \textcolor{black}{0.360} & \bf 0.430 & \textcolor{black}{\bf 0.444}\\  \hline

\hline
\end{tabular}}
\vspace{-1mm}
\caption{Ablation results on multimodal inputs and various feature extraction mechanisms. For the affect classification and quantification tasks, we report average scores.}
\label{tab:ablation}
\vspace{-5mm}
\end{table*}

%% file: table-comparison.tex
\begin{table*}[t]
\centering % centering table
\setlength{\extrarowheight}{5pt}
\resizebox{\textwidth}{!}{\begin{tabular}{l||c||c|c|c|c|c||c|c|c|c|c}
% \hline

\multirow{2}{*}{\bf Model} & \multirow{2}{6em}{\bf \centering Sentiment classification} & \multicolumn{5}{c||}{\bf Affect classification} & \multicolumn{5}{c}{\bf Affect quantification} \\\cline{3-7}\cline{8-12}

& & \bf Humor & \bf Sarcasm & \bf Offense & \bf Motivation & \bf Average & \bf Humor & \bf Sarcasm & \bf Offense & \bf Motivation & \bf Average \\ \hline

\hline
Baseline \cite{chhavi:semeval:2020:memotion} & 0.21765 & 0.51185 & 0.50635 & 0.49114 & 0.49148 & 0.50021 & 0.24838 & 0.24087 & 0.23019 & 0.48412 & 0.30089\\ \hline

Vkeswani IITK$^*$ & \textcolor{red}{\bf 0.35466}$^1$ & 0.47352 & 0.50855 & 0.49993 & 0.47379 & 0.48895 & 0.26171 & \textcolor{red}{\bf 0.25889}$^1$ & 0.26377$^2$ & 0.47379 & 0.31454 \\

George.Vlad Eduardgzaharia UPB$^*$ & 0.34539 & 0.51587$^3$ & \textcolor{red}{\bf 0.51590}$^1$ & \textcolor{red}{\bf 0.52250}$^2$ & 0.51909 & \textcolor{red}{\bf 0.51834}$^1$ & 0.24874 & 0.25392 & 0.24688 & 0.51909 & 0.31716$^3$ \\ 

Guoym$^*$ & 0.35197$^2$ & 0.51493 & 0.51099$^3$ & 0.51196 & 0.52065$^3$ & 0.51463$^2$ & \textcolor{red}{\bf 0.27069}$^1$ & 0.25028 & 0.25761 & 0.51126 & \textcolor{red}{\bf 0.32246}$^1$ \\

HonoMi Hitachi$^*$ & 0.34145 & 0.52136$^2$ & 0.44064 & 0.49116 & 0.51167 & 0.49121 & 0.26401$^3$ & 0.25378 & 0.24078 & 0.51679 & 0.31884$^2$ \\

Aihaihara$^{* \blacktriangle}$ & 0.35017$^3$ & - & - & - & - & - & - & - & - & - & - \\
Souvik Mishra Kraken$^*$ & 0.34627 & 0.51450 & 0.50415 & 0.51230 & 0.50708 & 0.50951$^3$ & 0.0 & 0.0 & 0.0 & 0.50708 & 0.12677 \\ \hline

Mayukh Memebusters$^*$ & 0.32540 & \textcolor{green!60!black}{\bf 0.52992}$^1$ & 0.48481 & \textcolor{green!60!black}{\bf 0.52907}$^1$ & 0.49069 & 0.50862 & 0.26127 & 0.23655 & \textcolor{red}{\bf 0.26512}$^1$ & 0.49069 & 0.31341 \\
Saradhix Fermi$^*$ & 0.24780 & 0.50214 & 0.49340 & 0.49648 & \textcolor{green!60!black}{\bf 0.53411}$^1$ & 0.50653 & 0.14053 & 0.23262 & 0.26141 & \textcolor{green!60!black}{\bf 0.53411}$^1$ & 0.29217 \\
Prhlt upv$^*$ & 0.33555 & 0.50956 & 0.51311$^2$ & 0.50556 & 0.50912 & 0.50934 & 0.25634 & 0.24382 & 0.24815 & 0.50912 & 0.31436 \\
Hg$^*$ & 0.32291 & 0.48583 & 0.50017 & 0.47254 & 0.52218$^2$ & 0.49518 & 0.21494 & 0.19354 & 0.23326 & 0.52218$^2$ & 0.29098 \\
Sourya Diptadas$^*$ & 0.34885 & 0.51387 & 0.49544 & 0.48635 & 0.49432 & 0.49750 & 0.26499$^2$ & 0.24498 & 0.24579 & 0.49432 & 0.31252 \\
Nowshed CSECU KDE MA$^*$ & 0.32301 & 0.49272 & 0.48705 & 0.50480 & 0.49053 & 0.49377 & 0.23701 & 0.25460$^2$ & 0.25172 & 0.50207 & 0.31135 \\

Jy930 Rippleai $^*$ & 0.33732 & 0.50035 & 0.48352 & 0.51589$^3$ & 0.52033 & 0.50502 & 0.25115 & 0.23783 & 0.25617 & 0.52033 & 0.31637 \\

Xiaoyu$^*$  & 0.34522 & 0.43376 & 0.44663 & 0.39965 & 0.48848 & 0.44213 & 0.25482 & 0.25415$^3$ & 0.24128 & 0.48848 & 0.30969 \\

Gundapu Sunil$^*$ & 0.33915 & 0.50156 & 0.49949 & 0.47850 & 0.49831 & 0.49446 & 0.23573 & 0.23011 & 0.26234$^3$ & 0.52132$^3$ & 0.31237  \\\hline

\hline
\name & \textcolor{green!60!black}{\bf 0.37621} & \textcolor{red}{\bf 0.52670} & \textcolor{green!60!black}{\bf 0.51936} & 0.51731 & \textcolor{red}{\bf 0.53102} & \textcolor{green!60!black}{\bf 0.52344} & \textcolor{green!60!black}{\bf 0.27134} & \textcolor{green!60!black}{\bf 0.26004} & \textcolor{green!60!black}{\bf 0.26821} & \textcolor{red}{\bf 0.53102} & \textcolor{green!60!black}{\bf 0.33265}\\ \hline

\hline
\end{tabular}}
\caption{Comparative study against baselines and various state-of-the-art systems. All scores are Macro-F1 as per the official evaluation metric of the `Memotion Analysis' shared task \cite{chhavi:semeval:2020:memotion}. Superscripts $^1$,$^2$, and $^3$ denote official rank of the system in the shared task. For each case, \textcolor{green!60!black}{\bf the best} and \textcolor{red}{\bf the second ranked} scores among all systems are highlighted in green!60!black and red texts, respectively. The first batch of results (after baseline) denotes a set of top three ranked systems for the three tasks (on average). \textbf{System$^*$:} Values taken from \citet{chhavi:semeval:2020:memotion}. $^{\blacktriangle}$ The model, called 
`Aihaihara' participated in only sentiment classification task.}
\label{tab:comparison}
\vspace{-5mm}
\end{table*}

%% file: table-multi-hops.tex
%%%% final version %%%%
\begin{table*}[!t]
\centering % centering table
\setlength{\extrarowheight}{3pt}
\resizebox{0.999\textwidth}{!}
{\begin{tabular}{l|l|c|c|c|c|c|c|c|c|c|c|c|c|c|c|c|c|c|c}
% \hline

\multirow{3}{*}{\bf Modality} & \multirow{3}{*}{\bf Hops} & \multicolumn{6}{c|}{\bf Sentiment classification} & \multicolumn{6}{c|}{\bf Affect classification} & \multicolumn{6}{c}{\bf Affect quantification} \\\cline{3-20}

& & \multicolumn{2}{c|}{\textcolor{black}{\bf D-Fusion}} & \multicolumn{2}{c|}{\bf AT-Fusion} & \multicolumn{2}{c|}{\bf ATMF} & \multicolumn{2}{c|}{\textcolor{black}{\bf D-Fusion}} & \multicolumn{2}{c|}{\bf AT-Fusion} & \multicolumn{2}{c|}{\bf ATMF} & \multicolumn{2}{c|}{\textcolor{black}{\bf D-Fusion}} & \multicolumn{2}{c|}{\bf AT-Fusion} & \multicolumn{2}{c}{\bf ATMF}\\ \cline{3-20} 

& & \textcolor{black}{\bf Test$_A$} & \textcolor{black}{\bf Test$_B$} & \bf Test$_A$ & \bf \textcolor{black}{Test$_B$} & \bf Test$_A$ & \bf \textcolor{black}{Test$_B$} & \textcolor{black}{\bf Test$_A$} & \textcolor{black}{\bf Test$_B$} & \bf Test$_A$ & \bf \textcolor{black}{Test$_B$} & \bf Test$_A$ & \bf \textcolor{black}{Test$_B$} & \textcolor{black}{\bf Test$_A$} & \textcolor{black}{\bf Test$_B$} & \bf Test$_A$ & \bf \textcolor{black}{Test$_B$} & \bf Test$_A$ & \bf \textcolor{black}{Test$_B$} \\ 

\hline

\multirow{2}{*}{BiLSTM - {\tt OCR} + VGG19} & Single-hop & \textcolor{black}{0.336} & \textcolor{black}{0.372} &  0.342 & \textcolor{black}{0.381} & 0.345 & \textcolor{black}{0.385} & \textcolor{black}{0.501} & \textcolor{black}{0.522} & 0.504 & \textcolor{black}{0.527} & 0.505 & \textcolor{black}{0.529} & \textcolor{black}{0.307} & \textcolor{black}{0.324} & 0.314 & \textcolor{black}{0.330} & 0.318 & \textcolor{black}{0.335}\\

& Multi-hop & \textcolor{black}{0.340} & \textcolor{black}{0.375} & 0.344 & \textcolor{black}{0.386} & 0.349 & \textcolor{black}{0.389} & \textcolor{black}{0.503} & \textcolor{black}{0.526} &  0.505 & \textcolor{black}{0.530} & 0.508 & \textcolor{black}{0.532} & \textcolor{black}{0.315} & \textcolor{black}{0.331} & 0.319 & \textcolor{black}{0.338} & 0.320 & \textcolor{black}{0.343}\\ \hline

\multirow{2}{*}{BiLSTM - {\tt OCR}$_{Seg}$ + VGG19} & Single-hop & \textcolor{black}{0.355} & \textcolor{black}{0.386} & 0.358 & \textcolor{black}{0.391} & 0.370 & \textcolor{black}{0.409} &  \textcolor{black}{0.510} & \textcolor{black}{0.528} & 0.512 & \textcolor{black}{0.533} & 0.517 & \textcolor{black}{0.540} & \textcolor{black}{0.322} & \textcolor{black}{0.344} & 0.326 & \textcolor{black}{0.348} & 0.329 & \textcolor{black}{0.352}\\ 

& Multi-hop & \textcolor{black}{0.364} & \textcolor{black}{0.405} & 0.372 & \textcolor{black}{0.413} & \bf 0.376 & \textcolor{black}{\bf 0.426} & \textcolor{black}{0.513} & \textcolor{black}{0.530} & 0.514 & \textcolor{black}{0.536} & \bf 0.523 & \textcolor{black}{\bf 0.545} & \textcolor{black}{0.324} & \textcolor{black}{0.346} & 0.327 & \textcolor{black}{0.351} & \bf 0.333 & \textcolor{black}{\bf 0.360} \\ \hline

\hline
\end{tabular}}
\vspace{-1mm}
\caption{Comparative study of different fusion mechanisms and effect of single-hop attention vs multi-hop attention.}\label{tab:ATMF}
\vspace{-6mm}
\end{table*}

%% file: figure-explainability.tex
\begin{figure}[t]
\centering
    \subfloat[\textcolor{black}{Input meme.}\label{fig:input_meme}]{
    \includegraphics[width=0.20\textwidth, height=0.20\textwidth]{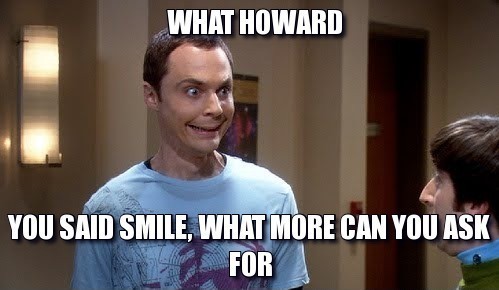}}
    \hspace{2em}
    \subfloat[\textcolor{black}{LIME output - image.}\label{fig:lime_image}]{
    \includegraphics[width=0.205\textwidth, height=0.205\textwidth]{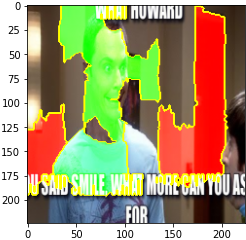}}
    \\
    \subfloat[\textcolor{black}{LIME output - text.}\label{fig:lime_txt}]{
    \includegraphics[width=\columnwidth]{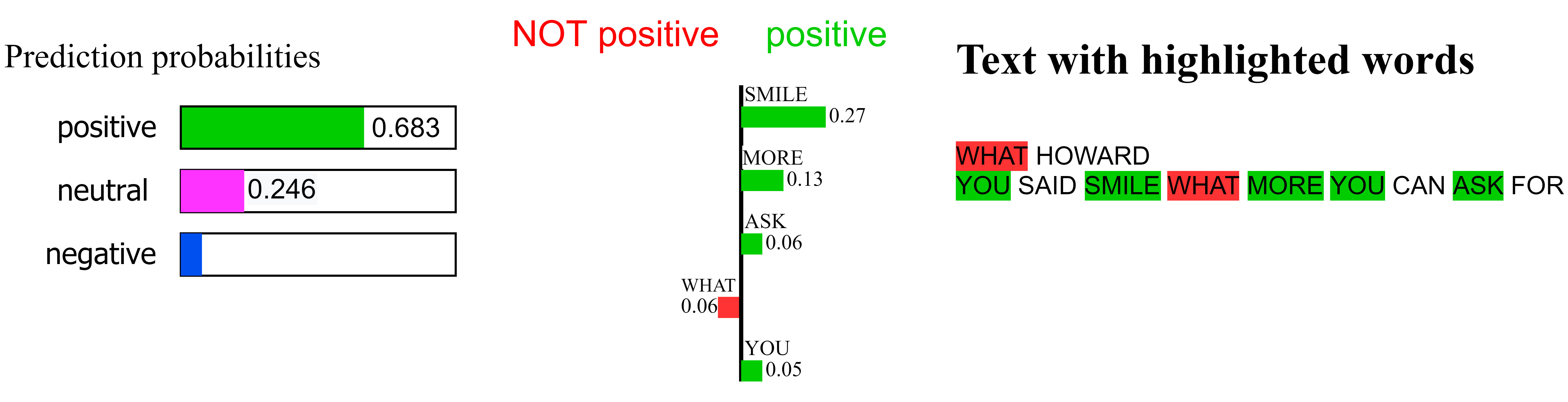}}
    \\
    % \hspace{2em}
    \subfloat[\textcolor{black}{Attention weights as computed by \name.} \label{fig:attention_text}]{
    \includegraphics[width=0.75\columnwidth]{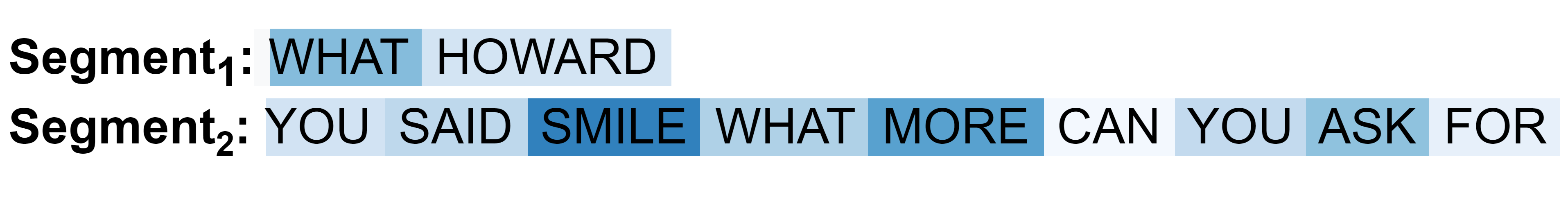}}\hspace{0.5em}
    % \hspace{0.5em}
    \caption{\textcolor{black}{Example of explanation by LIME on both visual and textual modalities and visualization of attention weights over text tokens obtained from \name.
    }}
    \label{fig:lime}
    \vspace{-5mm}
\end{figure}